# Vehicle Tracking Using Surveillance with Multimodal Data Fusion

Yue Zhang, Bin Song* *Member, IEEE,* Xiaojiang Du, *Senior Member, IEEE,* and Mohsen Guizani, *Fellow, IEEE*

*Abstract*—Vehicle location prediction or vehicle tracking is a significant topic within connected vehicles. This task, however, is difficult if only a single modal data is available, probably causing bias and impeding the accuracy. With the development of sensor networks in connected vehicles, multimodal data are becoming accessible. Therefore, we propose a framework for vehicle tracking with multimodal data fusion. Specifically, we fuse the results of two modalities, images and velocity, in our vehicle-tracking task. Images, being processed in the module of vehicle detection, provide direct information about the features of vehicles, whereas velocity estimation can further evaluate the possible location of the target vehicles, which reduces the number of features being compared, and decreases the time consumption and computational cost. Vehicle detection is designed with a color-faster R-CNN, which takes both the shape and color of the vehicles into consideration. Meanwhile, velocity estimation is through the Kalman filter, which is a classical method for tracking. Finally, a multimodal data fusion method is applied to integrate these outcomes so that vehicle-tracking tasks can be achieved. Experimental results suggest the efficiency of our methods, which can track vehicles using a series of surveillance cameras in urban areas.

I. INTRODUCTION

WITH technological advancements in vehicles and transportation system, motorists require comfort and intelligent driving, not only mobility. Thus, there has been a great deal of research which mainly falls into one of two directions. On one hand, researchers tend to develop more intelligent vehicles, or devices that can be attached to vehicles, bringing up several popular topics such as autonomous vehicles or driverless vehicles [1]. They intend to apply automatic control methods to build high functional vehicles so that drivers can be relieved from the stress and anxiety and enjoy experience as passengers would. The other set of studies focuses on establishing a whole system, namely connected vehicles, instead of emphasizing the functions of each individual vehicle.

Connected vehicles allows its agents to communicate and exchange data. With the popularity of the Internet of Things (IoT) [2] and device-to-device (D2D) [3] communications, the emergence of the Internet of Vehicles (IoV) [4] is not unrealistic. Basically, IoV contains vehicle-to-vehicle (V2V), vehicle-to-user (V2U), and vehicle-to-infrastructure (V2I) communications, according to different agents being considered. When the connection keeps growing, the vehicles may behave socially, referred to as Social Internet of Vehicles (SIoV) [5], or even establishing social vehicle swarms (SVS) [6]. With the ability to gather data from all available accesses, the decision is more comprehensive.

Specifically, this comprehension comes from two aspects. On one hand, data may be collected from similar sensors in different spatial or temporal relations, thus the blindness of a single sensor is avoided. On the other hand, data from dissimilar sensors can provide multimodal knowledge for a particular target so that decision makers can strategize optimally. For instance, color is one of the critical modalities in image recognition, without which one may fail to recognize objects. The former aspect is a fusion of quantity perspective. In other words, if the decision is unsatisfactory, the main reason is because of the deficiency of the sensors. Thus, we focus on the latter scenario, which fuses multimodal data for comprehensive or complete knowledge.

Multimodal data fusion [7] is one critical technology, especially in the era of big data [8], where data contain variety, velocity, and other characteristics. Similar to traditional data fusion, the key target is to handle the problems of automatic detection, association, and correlation [9], except that multimodality also expands data from several sources to several modalities.

One straightforward sensor network is surveillance systems, which are widely discussed by a number of data fusion methods [10]. Originally, object detection or tracking through surveillance videos is heavily dependent on the participation of humans, which introduces a great amount of energy and time consumption. Therefore, automatic detection and tracking have become popular topics, facilitating the areas of object detection and tracking in computer vision.

The primary goal in computer vision is to educate computers

This work has been supported by the National Natural Science Foundation of China (61372068, 61271173), and also supported by the ISN State Key Laboratory.

Y. Zhang and B. Song are with the State Key Laboratory of Integrated Services Networks, Xidian University, 710071, China (e-mail: y.zhang@stu.xidian.edu.cn, bsong@mail.xidian.edu.cn). Bin Song is the corresponding author.

X. Du is with Dept. of Computer and Information Sciences, Temple University, Philadelphia PA, 19122, USA (email: dxj@ieee.org)

M. Guizani is with Dept. of Electrical and Computer Engineering, University of Idaho, Moscow, ID 83844, USA (email: mguizani@gmail.com)

to see the natural world, as humans do. This, however, is a critical task for computers, since the inputs for computers are merely pixels or digits. Fortunately, with the assistance of deep learning [11], computer vision has been making remarkable progress towards a variety of tasks, such as recognition, detection, and tracking. Faster R-CNNs (region proposals with convolutional neural networks) [12] are one of the most efficient methods based on convolutional neuronal networks (CNNs), an analogy of vision systems of mammals.

When applying faster R-CNN directly to the vehicle-tracking problem, however, the result is not satisfactory. The main reason is because faster R-CNN only takes grey-scale pixels into consideration. Therefore, an improvement is to introduce multimodal deep learning [13], which gathers more comprehensive information in order to increase the accuracy of tracking.

Meanwhile, physical quantities, such as location, velocity, or acceleration, are also useful when tracking, which are all well applied in the Kalman filter [14]. The Kalman filter is capable of estimation and prediction based on previous data and new observations. Therefore, we apply the results of the Kalman estimation as one modality to locate the period when the target occurs. Specifically, in this paper, we propose a vehicle-tracking framework based on multimodal data fusion, synthesizing knowledge of grey-scale (such as shape and outline), color, and velocity features, as in Fig. 1, which might be the most significant ones when tracking vehicles through surveillance cameras. After these data are fused, a more accurate and efficient tracking result is achieved.

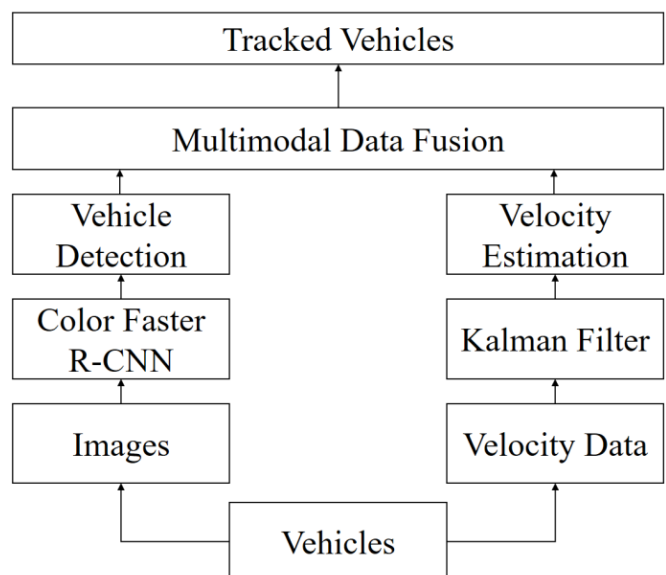

Fig. 1. The architecture of our proposed multimodal data fusion method for vehicle tracking. Two parallel modules exist in this framework. Images of vehicles, captured by cameras, are learned by the color-faster R-CNN method for vehicle detection purposes. The other module estimated the velocity using the Kalman filter based on the data obtained from motion sensors. Finally, these two modalities are fused by the multimodal data fusion method to achieve the vehicle-tracking task.

The main contributions of this paper are summarized as follows.
- We propose a novel multimodal data fusion framework for vehicle-tracking tasks, which improves the accuracy and efficiency more than that of the faster R-CNN;
- We add the velocity information, estimated by the Kalman filter, to the features of the faster R-CNN as an extra modality to reduce the number of candidates being compared;
- We have established an actual system with our proposed method, whose partial functions are already applicable in real tasks.

The rest of the paper is organized as follows. The next section will present some related work. The proposed methods are discussed in Section III and the corresponding experimental results are shown in Section IV. The final section concludes the whole paper.

## II. RELATED WORK

Most object tracking methods can be classified into two categories, discriminative or generative [15]. Discriminative methods firstly detect all objects in all images or videos, then re-identify them to project a trace within those belonging to the same object. Generative methods intend to build a model, containing abundant information about the object, so that the comparison is between the images and the model, which improves the accuracy of the matching. In this paper, we focus on discriminative methods.

From the perspective of discriminative object tracking [16], the intrinsic problem is converted into a classification or regression, which is widely discussed and studied in machine learning. It is highly related to recognition and detection, where the former emphasizes accuracy whereas the latter sacrifices accuracy to an acceptable degree for efficiency. Even though their targets are slightly different, both of them apply CNNs [17] to extract features from raw pixel inputs, which seems to be the most popular method.

Before CNNs became popular, other hand-designed features, such as SIFT [18] and HOG [19], were considered in object detection tasks. With the emergence of big data and powerful computational devices, such as GPUs and deep learning methods, especially CNNs in images, there has been an unstoppable trend in all related fields. As a result, many efficient object detection methods are proposed, one of which is R-CNN [20]. R-CNN applies a paradigm of regions [21] to select all candidate regions, then classifies the objects in each region. Later, a fast R-CNN [22] occurs and exceeds the velocity of the R-CNN until a faster R-CNN emerges [23]. Even though recent methods, namely YOLO [24], YOLO9000 [25], and mask R-CNN [26], claim to have a superior performance, we find that faster R-CNN is relatively stable. Therefore, we propose our method based on the faster R-CNN method.

After the detection process in the separated devices, connected vehicles provides a platform where knowledge can be shared for tracking tasks. This is a decision-making process, which relies on multimodal data. Intrinsically, multimodal data fusion is an integration and decision-making process with complicated inputs, thusly covering a variety of tasks such as detection and tracking [27]. Ideally, more modalities reduce the blindness of data, reaching a more comprehensive consideration and a wiser decision.

With the assistance of deep learning methods, connected vehicles is further improved and extended a closer relationship with other fields. One interesting perspective is treating vehicles as mobile components of home automation systems [28], whose ultimate goal is to increase the quality of the experience (QoE) of the residents. In the cyber layer, cloud computing [29], fog computing [30], and edge computing [31] are inseparable supportive technologies for data communication and computation.

Meanwhile, urban computing [32] intends to collect data and solve some actual urban problems, such as determining gas consumption or estimating air quality. Therefore, vehicles are natural data collection and execution devices in this framework. The final target agrees with the connected vehicles, IoV, and urban computing, which is the establishment of smart cities [33] to improve the quality of our lives.

## III. THE PROPOSED METHOD

Our proposed method fuses multimodal knowledge from image and velocity perspectives and contains two parallel modules, namely vehicle detection and velocity estimation. The former uses two features of target vehicles, gray-scale and colors, to provide candidates, and the latter estimates the possible period where the target occurs. After the multimodal information is fused, the outcomes should be more accurate than that of each individual one. We will discuss this process in detail.

### A. Faster R-CNN for Vehicle Tracking

As mentioned previously, we apply a discriminative method, namely faster R-CNN, to detect and track vehicles. Similar to other methods in supervised learning, the intrinsic problem is to minimize a cost function, as

$$L(\{p_i\},\{t_i\}) = \frac{1}{N_{cls}}\sum_i L_{cls}(p_i, p_i^*) + \frac{\lambda}{N_{reg}}\sum_i p_i^* L_{reg}(t_i, t_i^*). \quad (1)$$

Where $i$ is the index of an anchor and $p_i$ refers to the probability of detecting the anchor as an object. If the anchor is positive, the value of $p_i^*$ is 1 (0 otherwise), and the coordinates of the corresponding ground-truth box are $t_i^*$. $t_i$ is that of the predicted box. The whole process of faster R-CNN can be divided into a classification and a regression procedure, whose losses are measured by $L_{cls}$ and $L_{reg}$, respectively. $N_{cls}$ and $N_{reg}$ are normalization items, and the trade-offs are achieved by regularization parameter $\lambda$.

The application of a region proposal network (RPN) seems to be the main reason for the outstanding performance of faster R-CNN. The function of RPN is basically to estimate the location of objects in images, thusly reducing the influence of the background. Compared to fast R-CNN, the acceleration is achieved by sharing convolutional layers. This process is similar to the comparison between CNN and feed-forward neuron networks, where CNN greatly reduces the complexity by sharing the weight.

When only the faster R-CNN method is applied, the detection focuses on the appearance of vehicles and ignores other details, such as color. Thus, the only conclusion can be made on the types of vehicles, without revealing more specific information, as shown in the next section. Therefore, if faster R-CNN is applied directly to vehicle-tracking tasks, the results will not be acceptable. This motivates us towards multimodal data fusion to integrate more data to provide a comprehensive result.

### B. Color-Faster R-CNN

Since the results of applying the faster R-CNN are not satisfactory, we turn to discover more modalities, one of which is color information.

In original image recognition problems, extra information can be added either in series or in a parallel way. Faster R-CNN, however, behaves differently. Recall that faster R-CNN has two main processes, locating and then recognizing the objects. The process of locating objects relies on the classification, which is supervised instead of unsupervised. Thus, if a parallel framework is applied, color cannot be calculated since the algorithms have no target area. Therefore, we have to apply the color detection process after the location or objects have been found.

The color feature is extracted using a color histogram, which counts the distribution of images in three channels: hue, saturation, and values (HSV). Notice that even though other bases, such as RGB, are also available, our experiments suggest using HSV. The basic idea of the color histogram is to quantify the color space into an $s \times 1$ vector, and each pixel maps into the vector. Thus, this vector is a representation of the image, or some region of an image. In most scenarios, the color histogram is an efficient tool for distinguishing objects with different colors. For objects with similar colors, however, it shows limitations. Thus, the color histogram cannot be applied individually. Combined with faster R-CNN, color-faster R-CNN is able to classify objects in the dimensions of color and shape, which is critical for many applications, especially in vehicle tracking, where the data captured are limited to a few modalities.

### C. Velocity Estimation by Kalman Filter

Another modality we have applied in our method is velocity, which is estimated by the Kalman filter. In fact, the Kalman filter can handle a variety of parameters, such as position, velocity, or some angles. The reason for choosing velocity is that this quantity is our only accessible measurement by motion sensors, provided by the Henan Traffic Management Bureau. After acquiring the velocity, we can estimate the possible positions of our target vehicles. Thus, we can approximately estimate the indices of cameras and the period of its images being captured, based on the spatial relationships of the cameras.

The process of estimating the velocity of target vehicles can be achieved by the Kalman filter, which is popular in object tracking problems, especially in automatic control. Specifically, the prior state and its error covariance of the target vehicle at time $k$ are expressed as

$$\hat{X}_k = AX_{k-1} + \alpha_k, \quad (2)$$

$$\hat{P}_k = AP_{k-1}A^T + Q_k, \quad (3)$$

where $A$ is the transition matrix, which is designed by physical relationships. $\alpha_k$ is the uncertainty of the transition, which is usually assumed from a Gaussian distribution, $\alpha_k \sim \mathcal{N}(0, Q)$, where $Q$ is the covariance. $X$ is a column vector of size $n \times 1$, where each column is a measurement of the targets. Here, we simply apply two columns: the position and velocity of the vehicle(s) being tracked.

Then, we measure or observe the true state $X_k$ with
$$Z_k = BX_k + \beta_k, \quad (4)$$

where $B$ is the observation matrix, extracting the observable parameter of state vector $X$. Similarly, $\beta_k \sim \mathcal{N}(0, R)$ is the noise involved during the observation. Thus, the residual is obtained based on the difference between an actual measurement and a measurement prediction, as $(Z_k - B\hat{X}_k)$. Further, the systems will learn the optimal parameters for minimizing the residual, similar to the methods in supervised learning. After calculating the Kalman gain from (5), we are able to obtain the posteriori of the next state and covariance, as in (6) and (7).

$$K_k = \hat{P}_k H^T (H\hat{P}_k H^T + R)^{-1}, \quad (5)$$
$$X_{k+1} = \hat{X}_k + K(Z_k - B\hat{X}_k), \quad (6)$$
$$P_{k+1} = (I - K_k H_k)\hat{P}_k, \quad (7)$$

This prediction-correction process continues until the residual reduces to an acceptable level or until all test data are used.

In our model, velocities of vehicles can be captured through cameras. Thus, we apply these velocities as our actual observations to update our prediction model.

*D. Multimodal Data Fusion*

As the final process, we fuse both vehicle detection and velocity estimation for the vehicle-tracking task. Notice that more reasonable modalities can also be fused to improve the accuracy if necessary, even though we only present velocity. More modalities, however, may increase the burden of systems, whereas one less may decrease the accuracy. Therefore, choosing the modalities requires much attention.

The function of the velocities we applied is that they act as a filter to select important periods within all candidate features. Since each image of an object is converted into a feature vector through the CNNs, then all candidate features form a matrix $T$ of size $m \times n$, where $n$ is the number of candidates and $m$ is the length of each feature. Thus, the velocity filter is a sparse vector $F$ of size $n \times 1$, with $k$ ($k \ll n$) non-zero elements. The elements in filter $F$ are assumed to be of a normal distribution, as $F \sim \mathcal{N}(\mu, \sigma^2)$, where $\mu$ and $\sigma^2$ are related to the results $X$ and $P$ from the Kalman filter, respectively. Notice that even the density function should be greater than zero. We manually set a threshold to force most elements in $F$ to be zero. Then, we calculate the dot production to obtain the filtered set, whose elements are used for comparison rather than the whole set of $T$, as

$$C = F \odot T. \quad (8)$$

The obtained matrix, $C$, is a sparse version of the feature matrix,

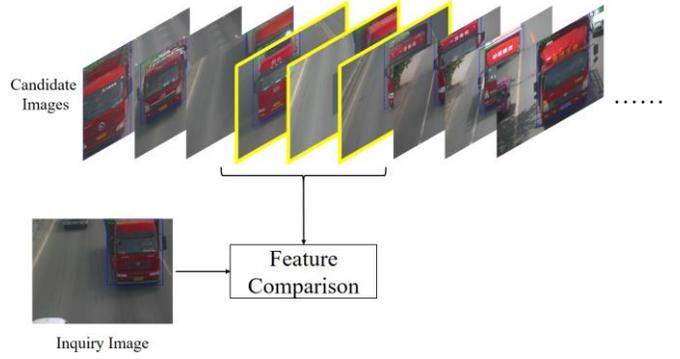

Fig. 2. The demonstration of applying the filter vector to candidate features. For a more direct demonstration, we have drawn the original images instead of its corresponding features. When the filter vector is applied, which is provided by the velocity estimation process, the number of candidate images or features are greatly reduced and only a few exist (highlighted by yellow edges). Thus, the feature matching or comparison process occurs only within the remainders, which significantly reduces the time consumption and computational cost.

where only $k$ non-zero features remain for comparison purposes, as demonstrated in Fig. 2. This will greatly reduce the time consumption and computational cost during the comparison process, since only $k$ objects remain to be compared.

In our simple model, only one extra modality is applied. Thus, we design a particular method for fusion. When a sophisticated scenario occurs, due to the data provided by sensors and requirements, a more general and robust fusion method is required. This may be difficult and may rely on promising topics across a variety of fields, such as the emergences of multimodal deep learning. Future work focuses on discovering the methods of multimodal deep learning and extending its capability to target problems with more intelligent requirements in connected vehicles.

IV. EXPERIMENTS

In this section, we present our simulation and experimental results. Firstly, we will present the unsatisfactory results when faster R-CNN is applied directly and the results of color are used faster R-CNN as a comparison. Then, the velocity estimation using the Kalman filter is shown. At last, we show the results of tracking the target vehicles. Most of our experiments are performed on an Ubuntu 14.04 LTS operating system (Intel@ Xeon(R) CPU E5-2630 0 @ 2.30GHz $\times$12, GeForce GTX 980Ti/PCle/SSE2).

*A. Vehicle Detection with faster R-CNN*

Firstly, we present the results of vehicle detection when faster R-CNN is directly applied without color information, as in Fig. 3. The time consumed by the training process is about 9 hours, with 80 thousand iterations. Our data consist of 4710 images, which are collected from actual surveillance cameras provided by the Henan Traffic Management Bureau.

Fig. 3 only presents four images out of all the detection results as a demonstration. Each object is labeled with a different color. The categories we apply are cars, trucks, buses, and motors, without special treatment for vans or other inter classes of vehicles. One interesting finding in panel (a) is the proof of the insensibility to color in the faster R-CNN, since both red and white trucks fall into the same class. Furthermore, this proof is solidified when two vehicles with similar colors are classified differently, as in panel (b). Therefore, if our intention is to track vehicles, color insensibility is not acceptable, since it brings much confusion to the results. Thus, a color-sensible detection is required, which motivates us to discover color-faster R-CNN, which is based on color features extracted by the color histogram method. We present the results of the color detection based on the categories of trucks, as in Fig. 4.

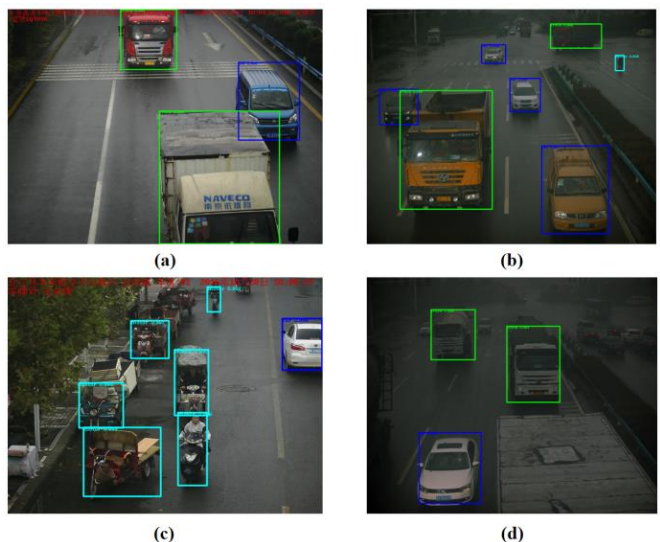

Fig. 3. Vehicle detection with faster R-CNN. Images from four different real surveillance cameras are presented. Different objects are indicated by colors. Notice that faster R-CNN is sensitive to the shapes of objects but not to color.

Panel (a) is the inquired image, which is manually inputted and expected to find the closest match. In fact, this mapping process can be finished with the original faster R-CNN method. Judging by the performance shown in Fig. 3, this may not be a wise choice. Images in panel (b) have the most similar results to the input in terms of color. Intuitively, all images contain a detected red truck, which is desired. Even though the results cannot guarantee to be the most similar in shape, they are still acceptable and valid since the comparison is based on the categories of trucks, which is already filtered by the faster R-CNN method.

This can also be regarded as an instance of application of multimodal data, even though shapes and colors are not technically two modalities. When color information is included, the results are more accurate and contain more detailed information. This is a scenario that a single type of data cannot achieve.

One problem is that we have to compare the input images with all candidates individually, which introduces a great cost especially when the set of images being compared is large. Meanwhile, shapes and colors cannot ensure that we have a single output, which is the vehicle most identical to the inquirer. Two opposite outcomes may occur here. One is that the output may fail to match the identical vehicles. This is probably caused by failures in selecting features. In other words, the shape or color may not be the optimal choice for detecting the vehicles for this task or for this database. On the other hand, when the output still contains a large number of images, the features may be deficient, which motives us to discover more features other than shape and color.

Based on these discussions, we tend to apply a very different modality, velocity, captured by different sensors. Combining the information of multiple modalities may discover a novel feature and perspective for real applications, such as more efficient features for vehicle detection and tracking to improve the accuracy, which will be further discussed in the next subsection.

### B. Velocity Estimation

Now, we estimate the velocity of the target vehicles based on the Kalman filter. The parameters in state vector $X$ are position and velocity, and only the velocity can be observed. The true velocity is assumed to be related to that of previous state, which is valid since the velocity changes continuously with little possibility of a sudden shifting. Ideally, the velocity can be estimated and observed at each state so that the system can efficiently learn the hidden true velocity. This, however, is highly unpractical. One straightforward reason is that our data are collected from surveillance systems which cannot record data when vehicles are beyond the sight of cameras. With limited data, the accuracy of detection cannot be ensured. Meanwhile, if the measurements and updates occur at each state, the computational cost is high. Therefore, we fix this problem by balancing these two extremes by mixing the truly observed data with artificial data, which are generated according to the previous state and noise. This is similar to the idea of semi-supervised learning, which combines labeled (observed) data with unlabeled ones.

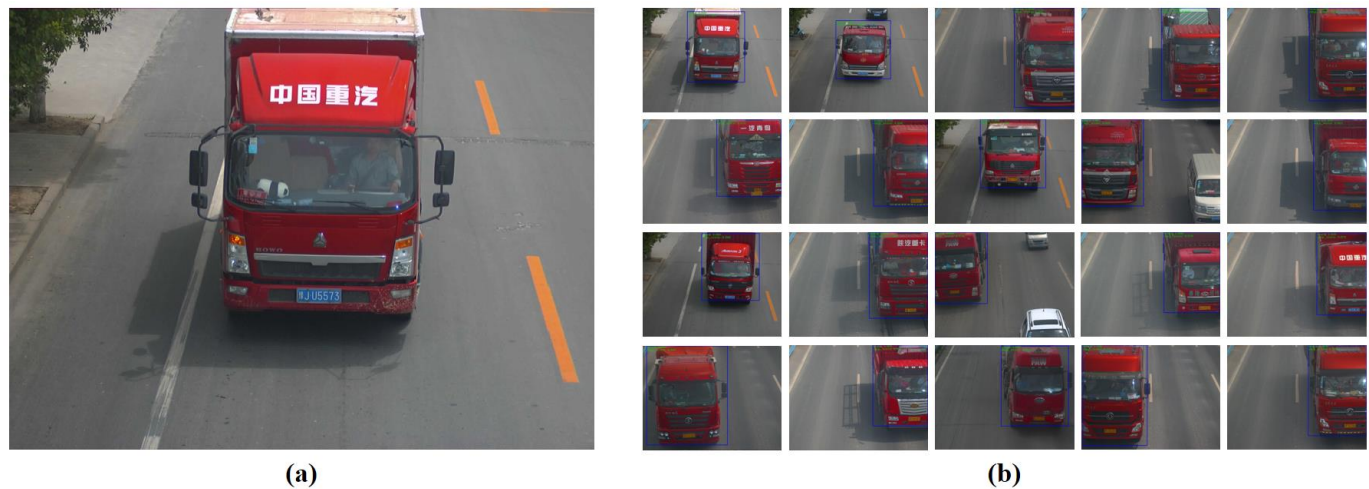

Fig. 4. A demonstration of the color-faster R-CNN method. When a target vehicle (panel (a)) is inquired, the top 20 similar vehicles are detected among all candidates, as in panel (b).

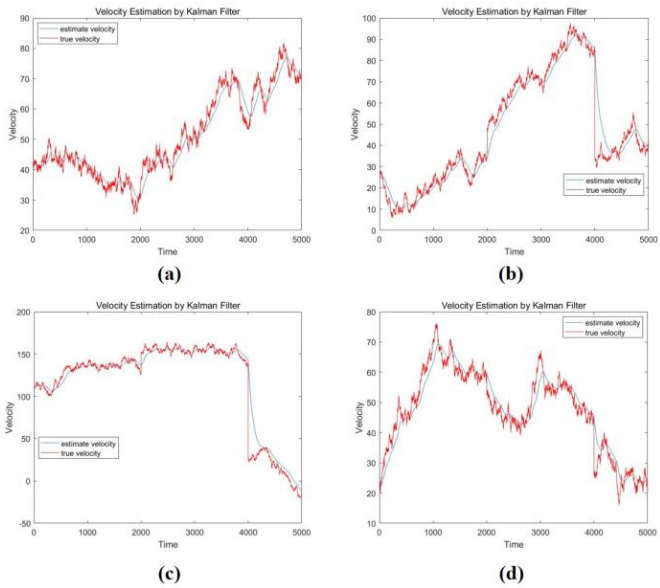

Fig. 5. Velocity estimation of target vehicles based on the Kalman filter. The values of the true observation (red lines) from surveillance cameras are at $t = [0, 2000, 4000]$, which can be successfully traced by the estimated values (blue lines).

Specifically, we generate the true velocity sequences individually and insert the observed values evenly into this sequence. When altering the variance of the generation process, the results are similar to a continuous outcome. Thus, we estimate the hidden sequence with the Kalman filter, as in Fig. 5.

Fig. 5 presents the estimation results from four instant vehicles with different variances of generating the hidden true values of velocities, according to the observed ones from motion sensors. It is straightforward that the estimated values can track the true ones, even if they change continuously. When the true values are forced against the current estimation, it shows a rapid change in the curves, as in panels (b) and (c), and the estimation can adapt to that change immediately. Therefore, with these estimated values, we can approximately predict the velocity anywhere. Furthermore, we apply the results so that they locate the appropriate cameras and the most probable period when the vehicle is across that camera, thusly reducing the computational cost and time spent on comparing the features of target vehicles and those of all vehicles captured by all surveillance cameras so that a more efficient vehicle tracking can be achieved.

### C. Vehicle Tracking

Finally, we present our vehicle-tracking results in an urban area in Zhengzhou, Henan province. Massive cameras exist so that we can apply the velocity estimations, discussed previously, to determine the possible cameras that capture the target vehicles and their corresponding times. The results are presented in Fig. 6, where only four vehicles (referred to with lowercase letters and colors), whose velocities are estimated in Fig. 5, and five cameras (referred to with uppercase letters) are shown. The numbers that are attached to letters is the order that the vehicles passed through these cameras. For instance, the path of red truck (a) is D-B-A, marked by a1-a2-a3. Thus, the trajectories of the target vehicles can be obtained. In fact, traditional object detection methods can also achieve a similar accuracy when they search and compare all images from all cameras. However, with multimodal data fusion, considering velocity as another modality in this task, we can select candidate cameras and reduce the number of cameras being compared, which greatly reduces the time consumption and the requirement of the hardware.

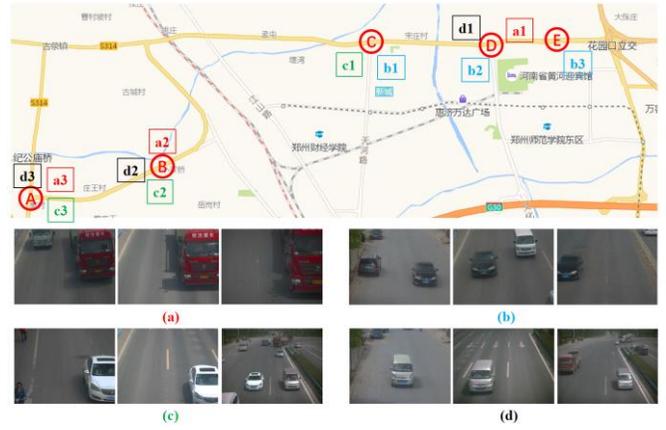

Fig. 6. Velocity tracking in an urban area. Only five cameras (marked by uppercase letters with circles) and four vehicles are presented as a demonstration, distinguished by letters as well as colors. The numbers attached to the letters on the map is the order of the corresponding vehicles passing through the location of the cameras.

Nevertheless, several problems still exist, one of which is that cameras cannot cover every area. The information from blind gaps between cameras is inaccessible, thusly we can only guess the states of vehicles there, which will have a negative effect on the accuracy. Another problem is the validation. Since we are using real data instead of existing data sets, we have no labels or prior knowledge of the data. Therefore, we apply the information from license plates which validate the identity of the vehicles. Furthermore, the velocity observed by motion sensors may not truly reveal, or even fail, to capture the velocity of target vehicles, which may affect the accuracy of the estimation.

Therefore, although some satisfactory results are presented, our framework is still an attempt for both vehicle tracking tasks and the multimodal data fusion method.

### V. CONCLUSION

In this paper, we have proposed a multimodal data fusion framework for vehicle tracking based on vehicle detection and velocity estimation. The vehicle detection process is achieved using the color-faster R-CNN, which is enhanced by the capability of extracting color information by the color histogram method, attached to the output of faster R-CNN. On the other hand, velocity, estimated by the Kalman filter, provides additional information for locating the possible time of appearance, thus the set being compared is shrunk, decreasing the time consumption and computational cost. The experimental results have shown that the color-faster R-CNN can detect vehicles accurately in terms of shapes and colors, which are the main evidences for human detection. After filtering the velocity modalities, the only limited features are compared with the inquiry images. The final trajectory of the target vehicles in the actual environments has also been presented. Our future work will focus on discovering sophisticated modalities for vehicle tracking and study advanced methods to fuse them in order to improve the capability and robustness of our model by discovering the potential of multimodal deep learning or applying deep learning methods for multimodal data fusion tasks. Meanwhile, improvements in the capability of sensors, such as providing more modalities, could lead to a more comprehensive and accurate outcome.

BIOGRAPHY

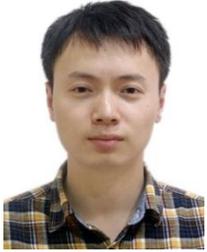
**Yue Zhang** received his B.S. degree in electronic information science and technology from the University of Electronic Science and Technology of China (UESTC), Chengdu, China in 2012 and his M.S. degree with Merit in computational intelligence from Sheffield University, Sheffield, U.K. in 2013. He is currently working towards his Ph.D. degree from Xidian University, Xi'an, China. His research interests include machine learning, multi-agent reinforcement learning, deep reinforcement learning, game theory, Internet of Things, and Big data.

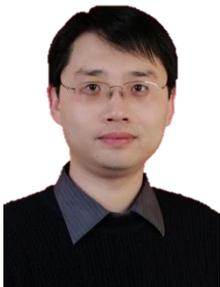
**Bin Song** received his BS, MS, and PhD in communication and information systems from Xidian University, Xi'an, China in 1996, 1999, and 2002, respectively. In 2002, he joined the School of Telecommunications Engineering at Xidian University where he is currently a professor of communications and information systems. He is also the associate director at the State Key Laboratory of Integrated Services Networks. He has authored over 50 journal papers or conference papers and 30 patents. His research interests and areas of publication include video compression and transmission technologies, video transcoding, error- and packet-loss-resilient video coding, distributed video coding, and video signal processing based on compressed sensing, big data, and multimedia communications.

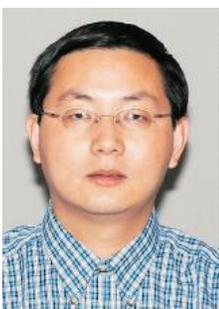
**Xiaojiang (James) Du** is a tenured associate professor in the Department of Computer and Information Sciences at Temple University, Philadelphia, USA. Dr. Du received his M.S. and Ph.D. degrees in electrical engineering from the University of Maryland College Park in 2002 and 2003, respectively. His research interests are wireless communications, wireless networks, security, and systems. He has authored over 170 journal and conference papers in these areas as well as a book, published by Springer. He won the best paper award at IEEE GLOBECOM 2014 and the best poster runner-up award at the ACM MobiHoc 2014. Dr. Du served as the lead Chair of the Communication and Information Security Symposium of the IEEE International Communication Conference (ICC) 2015 and a Co-Chair of Mobile and Wireless Networks Track of IEEE Wireless Communications and Networking Conference (WCNC) 2015. He is (was) a Technical Program Committee (TPC) member of several premier ACM/IEEE conferences. Dr. Du is a Senior Member of IEEE and a Life Member of ACM.

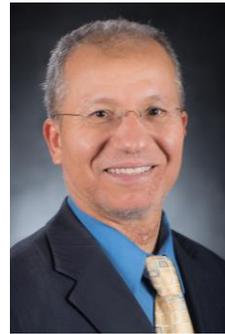
**Mohsen Guizani** (S'85–M'89–SM'99–F'09) received his bachelor's (with distinction) and master's degrees in electrical engineering and master's and doctorate degrees in computer engineering from Syracuse University, Syracuse, NY, USA in 1984, 1986, 1987, and 1990, respectively. He is currently a professor and the ECE Department chair at the University of Idaho. Previously, he served as the associate vice president of Graduate Studies, Qatar University, chair of the Computer Science Department, Western Michigan University, and chair of the Computer Science Department, University of West Florida. He also served in academic positions at the University of Missouri-Kansas City, University of Colorado-Boulder, Syracuse University, and Kuwait University. His research interests include wireless communications and mobile computing, computer networks, mobile cloud computing, security, and smart grid. He currently serves on the editorial boards of several international technical journals and is the founder and the editor-in-chief of the Wireless Communications and Mobile Computing journal (Wiley). He is the author of nine books and more than 400 publications in refereed journals and conferences. He guest edited a number of special issues in IEEE journals and magazines. He also served as a member, chair, and general chair at a number of international conferences. He was selected as the Best Teaching Assistant for two consecutive years at Syracuse University. He received the Best Research Award from three institutions. He was the chair of the IEEE Communications Society Wireless Technical Committee and the chair of the TAOS Technical Committee. He served as the IEEE Computer Society Distinguished Speaker from 2003 to 2005.